\documentclass{article-hermes}

\usepackage[latin1]{inputenc}
\usepackage[T1]{fontenc}

\usepackage{epsf}
\usepackage{hyperref}

\usepackage[french]{babel}

\usepackage{amsmath}
\usepackage{amsfonts}
\usepackage{amssymb}
\usepackage[ruled,vlined,french]{algorithm2e}
\usepackage{tipa}

\title[Distance d'{\'{e}}dition par blocs]{Algorithme de recherche approximative dans un dictionnaire fond{\'{e}} sur une distance d'{\'{e}}dition d{\'{e}}finie par blocs}

\author{Pascal Vaillant}

\address{Universit{\'{e}} Paris 13, Sorbonne Paris Cit{\'{e}}, LIMICS, (UMRS 1142)\\
          74 rue Marcel Cachin, 93017, Bobigny cedex, France\\
          INSERM, U1142, LIMICS, 75006, Paris, France\\
          Sorbonne Universit{\'{e}}s, UPMC Univ Paris 06, UMRS 1142, LIMICS, 75006, Paris, France\\
          \texttt{vaillant@univ-paris13.fr}}

\resume{Nous proposons un algorithme de recherche approximative de
  chaînes dans un dictionnaire à partir de formes
  altérées.  Cet algorithme est fondé sur une fonction de
  divergence entre chaînes~--- une sorte de distance
  d'{\'{e}}dition: il recherche des entrées pour lesquelles la
  distance à la chaîne cherchée est inférieure à un
  certain seuil.  La fonction utilisée n'est pas la distance
  d'édition classique (distance DL); elle est adaptée à un
  corpus, et se fonde sur la prise en compte de coûts
  d'altération élémentaires définis non pas sur des
  caractères, mais sur des sous-chaînes (des blocs de
  caractères).}

\abstract{We propose an algorithm for approximative dictionary lookup,
  where altered strings are matched against reference forms.  The
  algorithm makes use of a divergence function between strings---
  broadly belonging to the family of edit distances; it finds
  dictionary entries whose distance to the search string is below a
  certain threshold.  The divergence function is not the classical
  edit distance (DL distance); it is adaptable to a particular corpus,
  and is based on elementary alteration costs defined on character
  blocks, rather than on individual characters.}

\motscles{recherche approximative, correction d'erreurs, distance d'édition, fouille de textes}


\begin{document}

\maketitlepage


\section{Introduction: détection d'occurrences déformées de chaînes}
\label{intro}

La massification de l'accès aux technologies de l'information en
réseau, depuis une vingtaine d'années, a produit une croissance
exponentielle de sources de connaissance informelles,
hétérogènes, et non normalisées.  Ces sources contiennent
du texte non-normalisé car elles ont en commun le fait de ne pas
être produites en situation contrôlée, comme le sont les
sources qui passent par un cycle d'édition classique avant leur
mise à disposition du lecteur (rédaction dans l'objectif
spécifique d'une publication, relecture orientée, édition de
corrections, vérification finale).  Cette maigre caractéristique
commune, que l'on résume parfois en anglais par l'expression
générique «\,user generated content\,», ne constitue
qu'une définition en creux; et, ceci mis à part, on peut trouver
des types de texte de nature très différents.  Les sources de
leur «\,non-normalité\,» sont aussi diverses que les
pratiques qui les produisent.

Un premier type de corpus non-normalisé est ce que nous pouvons
appeler de manière générique le «\,forum
d'utilisateurs\,».  Plutôt que d'être produit par des acteurs
peu nombreux et détenteurs d'un contrôle institutionnel sur la
connaissance, ces corpus sont produits par une multitude d'acteurs
sans autorité~--- autre que celle, parfois, de leur expérience
personnelle subjective des sujets abordés.  Les sources de
connaissance non-normalisées de ce type intéressent les
chercheurs justement parce qu'elles reflètent non pas un état de
la connaissance, mais un état de la «\,réception\,» de la
connaissance par une communauté d'utilisateurs.  La moisson de
données dans ce type de corpus est centrale dans le domaine de la
fouille d'opinions.  Elle a d'autres usages plus spécifiques, comme
l'analyse de la compréhension qu'ont les patients des traitements
qui leur sont indiqués~\cite{HamonGagnayre2013}.  Un second type de
corpus non-normalisé est du type «\,notes d'expert\,».  Ils
sont produits par des acteurs qui peuvent être des autorités du
domaine, mais dans des conditions qui ne permettent pas à la
qualité rédactionnelle de faire partie des priorités du
processus de saisie de la connaissance. C'est le cas par exemple des
dossiers médicaux contenant des notes prises en texte libre par les
praticiens: ces notes utilisent souvent des abréviations et du
langage syntaxiquement et morphologiquement simplifié, et il faut
mettre en place des techniques spécifiques pour accéder à
leur contenu~\cite{CarrolKoelingPuri2012}.

Les corpus non-normalisés présentent des variations importantes
avec la langue standard, ce qui rend impossible d'utiliser directement
des outils de traitement automatique de la langue, même les plus
basiques~--- à commencer par la segmentation en mots ({\em
  tokenization}), qui nécessite de repérer des formes dans un
dictionnaire.

Pour résoudre ce problème, une première solution est de tenter
d'analyser le corpus non-normalisé comme une langue à part, et de
tirer d'une analyse distributionnelle les contours des différentes
variantes possibles d'une unité.  C'est l'approche proposée par
\cite{CarrolKoelingPuri2012}.  On renonce alors à la possibilité
d'utiliser des outils de TAL classiques (qui supposent l'accès aux
ressources linguistiques d'une langue normalisée), pour se concentrer
sur l'identification de termes par-delà la variété de leurs
manifestations concrètes, et sur la fouille de données que l'on peut
faire sur ces termes, leur distribution, et leurs co-occurrences.
L'inconvénient de cette approche est qu'elle nécessite un très grand
corpus d'apprentissage (à titre d'exemple, le travail cité se fonde
sur un corpus de 35 millions de mots), qui n'est pas toujours
disponible.

Une seconde solution consiste à normaliser le corpus avant de
l'analyser, quitte à perdre une partie de la connaissance dans le taux
d'oubli du processus de normalisation automatique.  C'est l'approche
illustrée par exemple par \cite{XueYinDavison2011} ou
\cite{PetzEtAl2012}.  Cette approche connaît elle aussi sa limite:
elle bute sur l'extrême variété des «\,déformations\,» qu'a pu subir la
chaîne de caractères correspondant à un mot normalisé lorsqu'elle se
manifeste dans des corpus non-normalisés.  C'est néanmoins dans ce
type d'approches que nous situons notre travail, car nous voulons
pouvoir reconnaître dans les textes non-normalisés des occurrences,
bien formées ou mal formées, de termes issus d'un thesaurus.

L'approche traditionnelle pour reconnaître des chaînes de
caractères ayant subi des «\,altérations\,» par rapport à
leur forme standard est de remplacer la recherche exacte dans un
dictionnaire par une recherche approximative.  Cette méthode repose
sur la possibilité de calculer une mesure de distance entre
chaînes de caractères, puis de fixer un seuil d'erreur
tolérée.  L'étalon de cette famille de distances est la {\em
  distance d'édition} de Damerau-Levenshtein, ou distance
DL~\cite{Damerau1964,Levenshtein1966}, qui compte le nombre
d'altérations élémentaires d'un caractère nécessaires
pour passer d'une chaîne à une autre.  De nombreux algorithmes
ont été proposés pour calculer efficacement une
variante\footnote{Les variantes possibles portent sur les
  altérations élémentaires considérées dans les
  diverses versions de la distance: insertion (I comme {\em
    insertion}), suppression (D comme {\em deletion}), inversion (R
  comme {\em reversal}) ou substitution (S comme {\em substitution}).
  Levenshtein prend en compte I,D et R; Damerau I, D, R et S; Wagner
  et Fischer I, D et S. Le principe reste identique.} de cette
distance d'édition: par exemple \cite{WagnerFischer1974} fournit
une version de référence utilisant la programmation dynamique
pour calculer la distance entre deux chaînes quelconques.
\cite{BaezaYatesNavarro1998,BaezaYatesNavarro1999} proposent
des algorithmes de recherche approximative de chaîne dans un texte;
et \cite{Bunke1994} est conçu spécifiquement pour comparer une
chaîne quelconque avec une chaîne présente dans un
dictionnaire.

Cependant, il a été remarqué que dans de nombreux domaines, la
distance d'édition classique ne permettait pas de prendre en compte le
fait que certaines substitutions ont une probabilité bien plus élevée
que d'autres, selon les contraintes du processus de génération de la
chaîne.  Plusieurs travaux ont donc cherché à «\,adapter\,» la
distance d'édition de façon à prendre en compte ces contraintes
spécifiques.  Le travail présenté ici se situe dans cette lignée; son
apport consiste à prendre en compte des coûts spécifiques, adaptés au
corpus, pour les opérations élémentaires, tout en incluant dans
celles-ci des substitutions de blocs au lieu de seules substitutions
de caractères.  Il présente un algorithme permettant d'optimiser la
recherche approximative de chaînes dans un dictionnaire en supposant
donnés ces paramètres de coût.  Il n'aborde pas la question de
l'apprentissage de ces paramètres sur des corpus annotés~--- cette
question est encore en cours d'exploration.

\section{La nécessité de distances d'édition adaptées}
\label{ndea}

Dans la suite, on notera $d(S,T)$, sans indice particulier, la
distance d'édition classique, dans sa variante utilisée par
\cite{WagnerFischer1974}, entre deux chaînes $S$ et $T$: il s'agit
du nombre minimum d'opérations élémentaires du type I
(insertion), D (suppression) et S (substitution) qui permettent de
passer de $S$ à $T$.

La décision de considérer une chaîne comme une occurrence
altérée d'une autre chaîne est fondée sur la
détermination d'un seuil de distance d'édition.  Ainsi, si l'on
considère la distance $d$, {\em servi} est une forme altérée
de {\em servie}, de distance 1 (une suppression). En fixant le seul à
1, on pourrait donc décider d'assimiler {\em servi} à {\em
  servie}, ce qui peut être utile en cas de faute de frappe ou de
faute d'orthographe.

Le handicap de la distance $d$ est qu'elle oblige à fixer un seuil
assez haut si l'on veut tolérer des altérations fréquentes dans les
textes non-normalisés.  Ainsi, pour qu'une recherche approximative
puisse reconnaître en {\em miolais} une forme altérée de la chaîne
{\em miaulait}, il est nécessaire de fixer un seuil de 3 (une
suppression, deux substitutions), ce qui laisse entrer dans les filets
de nombreuses chaînes qui seront jugées intuitivement moins
pertinentes dans le rôle de forme de référence candidate ({\em
  violais}, {\em musclais} ou {\em pilotais}).  En un mot, la distance
$d$ n'est pas capable de distinguer une séquence d'altérations
résultant d'un bruit aléatoire d'une séquence d'altérations
correspondant à une faute d'orthographe ou à une faute de frappe.

Si l'on veut une distance qui permet de corriger les erreurs
plausibles, il faut tenir compte des causes réelles des
altérations. Celles-ci dépendent du processus génératif, et dépendent
donc des moyens d'acquisition du corpus non-normalisé.  Nous citons
ici quelques exemples de causes d'altérations.

\begin{enumerate}

\item {\bf Diacritiques}: l'oubli ou l'omission de diacritiques est
  fréquent, notamment lorsque le dispositif technique utilisé
  pour saisir le texte oblige l'utilisateur à un effort
  supplémentaire pour entrer un caractère accentué
  (ex. utilisation d'un clavier QWERTY, ou de certains claviers
  virtuels où l'accès aux caractères accentués n'est pas
  direct). Exemple: {\em a cote} pour {\em à côté}.

\item {\bf Orthographe}: il s'agit des classiques erreurs de
  transcription phonème-graphème résultant d'une connaissance
  imparfaite du vocabulaire, ou d'une con\-science relâchée des
  règles d'orthographe grammaticale qui distinguent des formes
  homophones par des graphies spécifiques . Exemple: {\em on c'est
    mit à marché} pour {\em on s'est mis à marcher}.

\item {\bf Transcription}: lorsqu'un utilisateur écrit un mot dans
  le système orthographique d'une langue alors qu'il est plus
  habitué au système d'une autre langue, il peut utiliser
  involontairement des règles de transcription de phonèmes
  inadéquates.  Exemple: {\em lascia questo} écrit {\em lacha
    questo} par des individus parlant l'italien en famille mais
  scolarisés en français.

\item {\bf Clavier}: il s'agit des erreurs qui peuvent se produire
  lorsque le texte est saisi rapidement au clavier: frappe d'une
  touche voisine à la place, ou en plus, de la touche visée
  ({\em nises} pour {\em bises}, {\em avabnt} pour {\em avant}),
  saisie d'un caractère correspondant à la touche dans une autre
  configuration du clavier, à laquelle l'utilisateur est
  peut-être plus habitué ({\em ;ieux aue cq} pour {\em mieux que
    ça}), manque de synchronisation lors de l'utilisation des
  touches de fonction ou des touches mortes ({\em S7vres} au lieu de
  {\em Sèvres}, {\em äris} au lieu de {\em Paris}...),
  inversions dûes à une mauvaise synchronisation des deux mains
  ({\em slaut} pour {\em salut})...

\item {\bf Fréquence}: une partie des fautes de frappe ne sont pas
  liées à la disposition des touches du clavier, mais gardent un
  caractère prédictible: c'est celles qui se produisent
  lorsqu'un n-gramme très fréquent est machinalement tapé par
  l'utilisateur à la place d'un n-gramme moins fréquent ({\em
    guillement} pour {\em guillemet}).

\item {\bf Saisie semi-automatique}: les nouveaux procédés de
  saisie semi-automatique destinées à faciliter l'utilisation de
  dispositifs comme les claviers virtuels des {\em
    smart\-phones} ou des tablettes peuvent produire des erreurs
  lorsque l'utilisateur se repose trop sur eux sans vérifier leur
  résultat ({\em il est débauché} au lieu de {\em il est
    débouché}).

\item {\bf Reconnaissance optique de caractères}: certains textes
  sont acquis par une étape de numérisation d'un support
  imprimé, suivie d'une étape de reconnaissance optique de
  caractères.  Les logiciels de reconnaissance produisent des
  erreurs liées à la similarité de formes de certains
  caractères ({\em I}, {\em l}, et {\em 1}) ou groupes de
  caractères ({\em m} et {\em rn}).  Ces erreurs peuvent être
  réduites par l'utilisation de modèles de bigrammes, mais
  rarement complètement éliminées. Exemple: {\em carnées}
  pour {\em camées}.

\item {\bf Abréviations}: enfin, certains contextes (notamment la
  rédaction sous con\-trainte de temps) sont propices à
  l'utilisation fréquente d'abréviations qui sont
  compréhensibles pour un lecteur humain en contexte: {\em tt} pour
  {\em tout}, {\em pb} pour {\em problème}... Certaines de ces
  abréviations sont des acronymes, comme {\em RAS} pour {\em rien à
    signaler}; d'autres sont des dérivations phonographiques (ou
  «\,rébus typographiques\,»), fondées sur la manière
  dont se prononce un caractère dans sa forme isolée: du très
  répandu ({\em a+} pour {\em à plus}) au plus ludiques
  manifestations du «\,langage SMS\,» ({\em Lé13NRV});
  d'autres encore sont des dérivations idéographiques, comme
  l'usage de {\em ++} comme intensifieur (que l'on pourrait transcrire
  par {\em beaucoup}, {\em fort} ou {\em intense}, selon qu'il se
  situe dans une position qui implique un rôle d'adverbe ou
  d'adjectif).

\end{enumerate}

En considérant ces points, on constate la nécessité de pouvoir
utiliser une distance d'édition $d_C$ adaptée au corpus traité, qui ne
pénalise pas toutes les opérations élémentaires indifféremment.  Dans
chaque cas, on doit pouvoir calculer une distance qui impute une
pénalité plus faible aux altérations «\,probables\,» qu'aux
altérations aléatoires.  Ainsi, il s'agit de faire en sorte que
$d_C({\mbox{`o'}},{\mbox{`au'}})<2$ dans le cas d'un corpus contenant
de nombreuses fautes d'orthographe,
$d_C({\mbox{`ä'}},{\mbox{`Pa'}})<2$ dans le cas d'un corpus saisi au
clavier, $d_C({\mbox{`m'}},{\mbox{`rn'}})<2$ dans le cas d'un corpus
résultant d'une reconnaissance optique de caractères.

Il faut noter ici un élément important: pour pouvoir prendre en
compte une partie des cas de figure énumérés, il est
insuffisant de définir des coûts de transition sur des
opérations élémentaires touchant un caractère à la fois
(I, D, R ou S).  Il faut également pouvoir attribuer une
pénalité spécifique au remplacement du bloc `rn',
globalement, par le bloc `m'.  Nous devrons en tenir compte dans des
définitions qui vont suivre.

En outre, pour rendre la recherche approximative signifiante, il
convient que le seuil de distance considéré soit proportionnel à
la longueur de la chaîne (car l'ensemble des chaînes situés
sous un seuil de distance $d$ de 2 d'une chaîne de longueur 2
comporte l'ensemble de toutes les combinaisons possibles de deux
caractères).

\section{Travaux apparentés}
\label{edla}

Il existe une littérature abondante sur le problème de la recherche de
formes normalisées (ou correction d'erreurs).  Trois aspects sous
lesquels ce problème peut être considéré ont été tantôt abordés,
tantôt laissés de côté, par les différentes recherches publiées: celui
de l'optimisation des algorithmes de recherche lorsque la distance
d'édition doit être appliquée à la recherche approximative dans un
dictionnaire résultant du pré-traitement d'un corpus; celui de la
définition de distances d'édition adaptées; et celui de l'utilisation
de distances traitant des opérations de «\,blocs\,» de symboles, et
pas seulement de symboles isolés.

Le problème de la recherche approximative sur une distance
d'édition classique (du type distance DL) a été abordé par
\cite{WagnerFischer1974}, puis optimisé successivement par
\cite{MasekPaterson1980}, \cite{Bunke1994} et plus récemment par
\cite{CrochemoreLandauZiv2003}.  Cette dernière publication
constitue l'état de l'art, en termes d'optimisation, dans ce
domaine: elle utilise des fonctions de distance définies sur des
chaînes compressées et non plus sur les chaînes brutes;
cependant en amont seules les distances caractère par caractère
sont prises en compte.

\cite{Veronis1988} a décrit le problème des fautes de
transcription phonème-graphème: il a proposé un système
permettant de reconnaître des mots français mal
orthographiés, en considérant certaines paires de
sous-chaînes comme des classes d'équivalence, pour traiter de la
même manière les différentes transcriptions graphémiques
possibles d'un même phonème; cela le conduit par exemple à
attribuer le même coût à la transformation de {\em poteau} en
{\em ptautto} qu'à celle de {\em poteau} en {\em ptoteau}.  Il note
par ailleurs que le choix d'attribuer un coût nul à ces segments
équivalents n'est qu'une décision arbitraire, et que l'on peut
étendre le principe à l'attribution de coûts non-nuls, mais
non uniformes, fondés sur les fréquences d'erreur.  Cependant,
il ne donne pas de cadre formel de définition de la distance sur
laquelle est fondée son algorithme, et l'absence de propriété
de séparabilité implique que l'on ne peut pas borner, en
théorie, la complexité de l'implémentation.

Certains auteurs ont abordé le problème en utilisant des
distances qui s'éloignent nettement du principe de la distance
d'édition.  Ainsi \cite{PollockZamora1984} ont réalisé un
algorithme qui corrige automatiquement un grand nombre de fautes de
frappe en se fondant sur un corpus d'erreurs relevés dans les
premières versions d'articles scientifiques; cependant leur calcul
ne se fonde pas sur une distance d'édition, mais sur la comparaison
de «\,clés\,» obtenues à partir des chaînes en listant
leur consonnes distinctes, puis leurs voyelles distinctes, par ordre
d'apparition.  \cite{Ukkonen1992} a utilisé une distance fondée
sur le profil de distribution de $q$-grammes distincts de l'alphabet
dans les chaînes comparées.

Une autre famille de travaux a exploré les complexités combinatoires
du calcul de distances d'édition permettant des opérations sur des
blocs de caractères (insertions, suppressions, déplacements ou
remplacements de blocs entiers).  Les conclusions de ces travaux sont
assez pessimistes sur la complexité de sous-problèmes trop peu
restreints de ce problème général.  Ainsi, \cite{ShapiraStorer2003} et
\cite{LoprestiTomkins1997} montrent que dans le cas général, ce
problème est NP-complet; cependant, \cite{ShapiraStorer2011} comme
\cite{LoprestiTomkins1997} fournissent des exemples de limitations
imposées aux d'opérations de blocs, qui permettent de borner le
problème.  Ces travaux fournissent des algorithmes de comparaison de
séquences, mais celles-ci ne sont pas spécifiquement adaptées à la
recherche dans un dictionnaire.

Notre travail se situe à l'intersection de ces trois lignées,
puisqu'il consiste à définir un algorithme de recherche
approximative de chaîne dans un dictionnaire, qui est en même
temps fondé sur l'évaluation d'une distance d'édition
adaptée à un corpus donné (généralisable à tous
types de «\,sources d'altération\,» du texte: erreurs de
transcription des phonèmes, fautes de frappe, erreurs d'OCR...)

\section{Distance utilisée}
\label{distance}

Nous reprenons tout d'abord une définition formelle classique de la
distance d'édition, telle qu'elle est donnée par exemple par
\cite{Mohri2003}, puis l'étendons pour prendre en compte des
altérations de blocs.

Soit $\Sigma{}$ un alphabet composé d'un nombre fini de symboles
distincts, et $\Sigma{}^*$ le monoïde libre engendré par $\Sigma{}$
(ensemble de chaînes constituées par la concaténation de symboles de
$\Sigma{}$), muni d'un élément neutre $\epsilon{}$ (chaîne vide) pour
l'opération de concaténation.  On appelle $\Omega{}$ l'ensemble des
opérations élémentaires d'édition d'un symbole, que l'on peut définir
comme le produit cartésien de $\Sigma{}$ par lui-même:
$\Omega{}=\Sigma{}\cup{}\{\epsilon{}\}\times{}\Sigma{}\cup{}\{\epsilon{}\}-\{(\epsilon{},\epsilon{})\}$. Le
monoïde libre $\Omega{}^*$ engendré par $\Omega{}$ peut être considéré
comme l'ensemble des séquences d'opérations élémentaires qui
permettent de passer d'une chaîne à une autre; on l'appelle l'ensemble
des {\em alignements} de deux chaînes de $\Sigma{}^*$.  On note $h$ le
morphisme de $\Omega{}^*$ vers le produit cartésien
$\Sigma{}^*\times{}\Sigma{}^*$, et l'on peut noter par exemple que si
$\omega{}=(a,\epsilon{})(b,\epsilon{})(a,b)(\epsilon{},b)$, alors
$h({\omega{}})=(aba,bb)$ (l'un des moyens de passer de `aba' à `bb'
est de détruire un `a', de détruire un `b', de remplacer un `a' par un
`b', puis d'insérer un `b').

Soit $c$ une fonction de coût sur l'ensemble des opérations
élémentaires d'édition de symbole: $c:
\Omega{}\rightarrow{}\mathbb{R}^{+}$.  On définit le coût d'un
alignement comme la somme des coûts des paires élémentaires qui le
constituent: pour $\omega{}=\omega{}_{0}\dots{}\omega{}_{n} \in{}
\Omega{}^*$, $c(\omega{})=\sum_{i=0}^{n}{c(\omega{}_{i})}$. La {\em
  distance d'édition} entre deux chaînes $x\in\Sigma{}^*$ et
$y\in\Sigma{}^*$ est alors définie comme l'alignement de coût minimal
entre $x$ et $y$: $d_c(x,y)=\min{}\{c(\omega{}):h(\omega{})=(x,y)\}$.
\cite{Mohri2003} montre ensuite que $d_c$ vérifie les axiomes qui
définissent une distance.

On voit que cette définition classique de la distance d'édition
ne considère que des transitions élémentaires de type S, I ou
D (les éléments de $\Omega{}$ sont des substitutions lorsqu'il
s'agit de couples de symboles élémentaires, des insertions si le
premier terme du couple est $\epsilon{}$, et des suppressions si le
deuxième terme du couple est $\epsilon{}$).  Or nous avons expliqué
plus haut l'utilité d'une distance qui puisse prendre en compte des
altérations portant sur des blocs.

Nous proposons, pour cela, une nouvelle définition exposée dans
les paragraphes suivants.  Dans la suite, on notera les chaînes par
des majuscules latines ($S$), les symboles (variables) par des
minuscules grecques ($\sigma{}$), les symboles (valeurs) par des
minuscules latines (a); les chaînes constituées d'une
séquence de symboles seront entourées par des guillemets anglais
(ce qui permettra de distinguer la chaîne ``a'' du symbole a); la
concaténation sera notée explicitement par le symbole $+$, et
$|S|$ désignera la longueur d'une chaîne $S$.

On note $\Gamma{}^0$ l'ensemble des «\,courts-circuits\,» entre
blocs. $\Gamma{}^0$ est un sous-ensemble fini de
$\Sigma{}^*\times{}\Sigma{}^*-(\epsilon{},\epsilon{})$ possédant
les propriétés suivantes:

\noindent
\begin{math}
\forall{}\sigma{}\in{}\Sigma{}, ({\mbox{``}}\sigma{}{\mbox{''}},{\mbox{``}}\sigma{}{\mbox{''}})\in{}\Gamma{}^0\\
\forall{}(G,H)\in{}\Gamma{}^0, (H,G)\in{}\Gamma{}^0\\
\forall{}(G,H)\in{}\Gamma{}^0, |G|>1 \Rightarrow{} H\neq{}G\\
\forall{}(G,H)\in{}\Gamma{}^0, |H|>1 \Rightarrow{} H\neq{}G
\end{math}

L'ensemble $\Gamma{}^0$ contient une liste de couples de blocs entre
lesquels il y a confusion possible. Il contient, trivialement,
l'ensemble des couples de chaînes contenant un seul caractère,
lorsque ce caractère est identique dans les deux éléments du
couple.  On peut y inclure par ailleurs des couples de chaînes d'un
seul caractère pour des caractères entre lesquels la confusion
est fréquente (``è'' et ``ê''), ou des couples de blocs
(``m'' et ``rn'').  On n'y inclut pas des couples de blocs identiques
de plus d'un caractère (comme (``ab'',``ab'')), car cette
information serait redondante avec la présence des couples
identiques d'un seul caractère de longueur ((``a'',``a'') et
(``b'',``b'')).  Cet ensemble $\Gamma{}^0$ est invariant par
symétrie diagonale sur $\Sigma{}^*\times{}\Sigma{}^*$.

On dispose d'une fonction de coût
$c^0:\Gamma{}^0\rightarrow{}\mathbb{R}^+$ possédant ces propriétés:

\noindent
\begin{math}
\forall{}\sigma{}\in{}\Sigma{}, c^0({\mbox{``}}\sigma{}{\mbox{''}},{\mbox{``}}\sigma{}{\mbox{''}})=0\\
\forall{}(G,H)\in{}\Gamma{}^0, c^0(G,H)=c^0(H,G)\\
\forall{}(G,H)\in{}\Gamma{}^0, G\neq{}H \Rightarrow{} c^0(G,H)>0\\
\forall{}(G,H)\in{}\Gamma{}^0, c^0(G,H)<\max{}(|G|,|H|)
\end{math}

La troisième condition permet de construire, sur la base de
$(\Gamma{}^0,c^0)$, une véritable divergence, qui permette de
séparer systématiquement deux chaînes non-identiques.  Le
choix fait par \cite{Veronis1988} de traiter «\,peautto\,»
et «\,poteau\,» comme strictement équivalents ne nous semble
pas se justifier.

La quatrième condition assure que $\Gamma{}^0$ et $c^0$ n'encodent pas
d'information inutile: la borne supérieure signifie que le coût de la
transition dans un court-circuit doit être strictement inférieur au
coût du remplacement mécanique, symbole par symbole, de l'une des deux
chaînes par l'autre.

On note $\Gamma{}^1$ l'extension de $\Gamma{}^0$ qui contient en plus
toute la liste des couples possibles de blocs d'un seul symbole:

\noindent
\begin{math}
\forall{}\sigma{}\in{}\Sigma{}, ({\mbox{``}}\sigma{}{\mbox{''}},\epsilon{})\in{}\Gamma{}^1\\
\forall{}\sigma{}\in{}\Sigma{}, (\epsilon{},{\mbox{``}}\sigma{}{\mbox{''}})\in{}\Gamma{}^1\\
\forall{}\sigma{},\tau{}\in{}\Sigma{}, ({\mbox{``}}\sigma{}{\mbox{''}},{\mbox{``}}\tau{}{\mbox{''}})\in{}\Gamma{}^1\\
\forall{}(G,H)\in{}\Gamma{}^0, (G,H)\in{}\Gamma{}^1\\
\forall{}(G,H)\in{}\Gamma{}^1, |G|>1 \Rightarrow{} (G,H)\in{}\Gamma{}^0\\
\forall{}(G,H)\in{}\Gamma{}^1, |H|>1 \Rightarrow{} (G,H)\in{}\Gamma{}^0
\end{math}

Cette extension permet de couvrir l'intégralité des
transformations possibles d'une chaîne en une autre ($\Omega{}$) en
incluant toutes les opérations élémentaires D, I et S pour
chaque paire de symboles de $\Sigma{}$.

On dispose d'une fonction de coût
$c^1:\Gamma{}^1\rightarrow{}\mathbb{R}^+$ définie de la manière
suivante:

\begin{math}
\forall{}(G,H)\in{}\Gamma{}^1, \left\{ \begin{array}{ll}
c^1(G,H) = c^0(G,H) & {\mbox{si }}(G,H)\in{}\Gamma{}^0\\
c^1(G,H) = \max{}(|G|,|H|) = 1 & {\mbox{sinon}}
\end{array} \right.
\end{math}

La fonction de coût $c^1$ est un prolongement de $c^0$ qui permet
d'intégrer les coûts par défaut des opérations I, D et S
pour les couples qui ne sont pas des courts-circuits.  La
définition de $\Gamma{}^1$ garantit que dans ce cas
$\max{}(|G|,|H|)=1$.

Ces bases étant posées, on peut à présent définir une fonction de coût
$d$ sur $\Sigma{}^*\times{}\Sigma{}^*$ par les propriétés suivantes:

Soit $\mathcal{S}(S)$ l'ensemble des segmentations de $S$ en blocs
(sous-chaînes) plus petits, c'est-à-dire l'ensemble des suites
finies $(S_0,\dots{},S_n)$ de chaînes $S_k$ (possiblement vides)
telles que $S=S_0+\dots{}+S_n$ (où $n\in{}\mathbb{N}$).

On définit $d$ pour tout couple de chaînes
$(S,T)\in{}\Sigma{}^*\times{}\Sigma{}^*$ par:

\begin{math}
d(S,T) = \min_{\substack{n\leqslant{}\max{}(|S|,|T|)\\ (S_0,\dots{},S_n)\in{}\mathcal{S}(S)\\ (T_0,\dots{},T_n)\in{}\mathcal{S}(T)}}(\Sigma{}_{i=0}^{n}{c^1(S_i,T_i)})
\end{math}

Cette fonction est définie et il s'agit d'une divergence
symétrique, ce qui peut se démontrer comme indiqué dans les
paragraphes ci-dessous.

\noindent
{\bf Existence:} par définition du monoïde libre $\Sigma{}^*$, toute
chaîne peut être obtenue par concaténation de chaînes composées d'un
seul caractère: si
$S={\mbox{``}}\sigma{}_0\dots{}\sigma{}_{n}{\mbox{''}}$,
$S={\mbox{``}}\sigma{}_0{\mbox{''}}+\dots{}+{\mbox{``}}\sigma{}_{n}{\mbox{''}}$.

Soit un couple de chaînes quelconque $(S,T)$ de
$\Sigma{}^*\times{}\Sigma{}^*$; on supposera pour le raisonnement que
$S$ désigne la plus longue des deux chaînes ($|S|=n$, $|T|=m$,
$n\geqslant{}m$). On peut décomposer $S$ et $T$ en segments de
longueur 1:
$S={\mbox{``}}\sigma{}_0{\mbox{''}}+\dots{}+{\mbox{``}}\sigma{}_{n}{\mbox{''}}$,
$T={\mbox{``}}\tau{}_0{\mbox{''}}+\dots{}+{\mbox{``}}\tau{}_{m}{\mbox{''}}$.

Si $n=m$, il existe au moins un couple de segmentations de
$\mathcal{S}(S)\times{}\mathcal{S}(T)$ permettant de définir une
transformation de $S$ en $T$ (il s'agit de la série de $n$
substitutions d'un seul symbole à la fois): $\forall{}i\in{}[0,n]$,
$({\mbox{``}}\sigma{}_i{\mbox{''}},{\mbox{``}}\tau{}_i{\mbox{''}})\in{}\Gamma{}^1$,
et
$c^1({\mbox{``}}\sigma{}_i{\mbox{''}},{\mbox{``}}\tau{}_i{\mbox{''}})$
est définie. Si $n>m$, il existe là encore au moins un couple de
segmentations permettant de définir une transformation de $S$ en $T$
(une série de substitutions de $m$ caractères suivie d'une série
d'insertions de $n-m$ caractères): $\forall{}i\in{}[0,m]$,
$({\mbox{``}}\sigma{}_i{\mbox{''}},{\mbox{``}}\tau{}_i{\mbox{''}})\in{}\Gamma{}^1$,
et
$c^1({\mbox{``}}\sigma{}_i{\mbox{''}},{\mbox{``}}\tau{}_i{\mbox{''}})$
est définie, et $\forall{}i\in{}]m,n]$,
    $({\mbox{``}}\sigma{}_i{\mbox{''}},\epsilon{})\in{}\Gamma{}^1$, et
    $c^1({\mbox{``}}\sigma{}_i{\mbox{''}},\epsilon{})$ est définie.

Dans tous les cas il existe donc au moins un élément de
$\mathcal{S}(S)$ et un élément de $\mathcal{S}(T)$ pour lesquels le
calcul de $c(S,T)$ est possible.

\noindent
{\bf Unicité:} garantie par la définition du minimum sur un
sous-ensemble de $\mathbb{R}^+$.

\noindent
{\bf Axiome de séparation:} On note tout d'abord trivialement que
$\forall{}S\in{}\Sigma{}^*, d(S,S)=0$: en effet la transformation de
$S$ en elle-même, définie par la transformation (à coût
nul) de chacun de ses symboles par le symbole identique, a un coût
total nul:
$\Sigma{}_{i=0}^{n}c^1({\mbox{``}}\sigma{}_{i}{\mbox{''}},{\mbox{``}}\sigma{}_{i}{\mbox{''}})=0$

Il s'agit ensuite de montrer que deux chaînes différentes ne
peuvent avoir une distance nulle: $\forall{}S,T\in{}\Sigma{}^*,
d(S,T)=0 \Rightarrow{} S=T$. Supposons deux chaînes $S$ et $T$
telles que $d(S,T)=0$. Il existe au moins un entier
$n\leqslant{}\max{}(|S|,|T|)$, une segmentation $(S_0\dots{}S_n)$ de
$S$ et une segmentation $(T_0\dots{}T_n)$ de $T$, telles que
$\forall{}i\in{}[0,n]$, $c^1(S_i,T_i)=0$ (les coûts ne pouvant
être négatifs, pour que la somme soit nulle, il faut que chaque
terme soit nul). Par définition de $c^1$, $c^1(S_i,T_i)=0
\Rightarrow{} (S_i,T_i)\in{}\Gamma{}^0{\mbox{ et
}}c^1(S_i,T_i)=c^0(S_i,T_i)$. Par définition de $c^0$,
$c^0(S_i,T_i)=0 \Rightarrow{} S_i=T_i$. Donc $\forall{}i\in{}[0,n]$,
$S_i=T_i$. Donc $S=S_0+\dots{}+S_n=T_0+\dots{}+T_n=T$.

\noindent
{\bf Symétrie:} $c^0$ étant symétrique, $c^1$ est
symétrique; $c^1$ étant symétrique, $d$ est symétrique: en
effet
$\forall{}(S_0\dots{}S_n)\in{}\mathcal{S}(S),(T_0\dots{}T_n)\in{}\mathcal{S}(T)$,
$\Sigma{}_{i=0}^{n}{c^1(S_i,T_i)}=\Sigma{}_{i=0}^{n}{c^1(T_i,S_i)}$.

En revanche, la fonction $d$ n'est pas véritablement une distance
métrique car elle ne vérifie pas la propriété
d'inégalité triangulaire.  Cependant, cette propriété
n'est pas nécessaire pour l'algorithme décrit ci-dessous.

Il est important de noter que la définition de $d$ repose sur
l'établissement préalable de $\Gamma{}^0$ et de $c^0$, qui sont
des données que l'on doit établir par rapport aux besoins du
corpus à traiter, afin de prendre en compte la réalité des
types d'erreur possibles selon le contexte (cf. plus haut, section
\ref{ndea}).  Il ne s'agit donc pas d'une distance fondée sur des
clés d'index comme dans \cite{PollockZamora1984}, ou d'une distance
d'édition étendue ou généralisée comme celle
proposée dans \cite{FuadMarteau2008}, mais d'une distance
adaptée au corpus à traiter; nous la notons donc dans la suite
$d_C$.

\section{Algorithme}
\label{algo}

\subsection{Définition}

L'algorithme que nous proposons est adapté aux cas de figures où
la recherche de chaîne doit se faire dans un index qui a été
pré-compilé sous forme d'arbre lexicographique (``{\em trie}'').
\cite{BaezaYatesNavarro1998} traitent du cas où la recherche
approximative de chaînes se fait dans un grand corpus non
indexé, même s'il est appelé «\,dictionnaire\,» par les
auteurs; \cite{Bunke1994} traite du cas où la recherche se fait
dans un automate à états finis stockant l'une des deux
chaînes en tant qu'entrée de dictionnaire, mais se limite à
la distance d'édition par symboles, et non par blocs.

Le problème de la recherche d'une chaîne de longueur $n$ dans un
ensemble de $k$ mots de longueur $m$ est, dans le cas le plus général,
de complexité $O(n\times{}m\times{}k)$ (algorithme naïf) si le corpus
n'est pas indexé.  Il peut être amélioré par un pré-traitement de la
chaîne de longueur $n$ (par des algorithmes comme ceux de Knuth,
Morris et Pratt ou Boyer-Moore), mais le temps de recherche reste
dépendant de la taille du corpus.

\begin{figure}[hbtp]
\begin{center}
\epsfxsize=50mm
\mbox{\epsffile{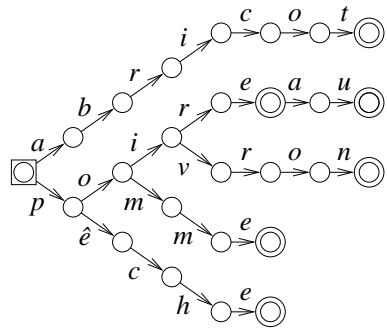}}
\caption[Recherche dans un arbre lexicographique]{\small Recherche
  exacte d'une chaîne dans un arbre lexicographique encodant le
  dictionnaire
  $\{$``abricot'',``poire'',``poireau'',``poivron'',``pomme'',``pêche''$\}$:
  on parcourt l'arbre depuis le n{\oe{}}ud de départ jusqu'au moment
  où on ne trouve plus d'arc sortant correspondant au symbole suivant
  dans la chaîne recherchée. Pour une chaîne $S$ de longueur $n$, le
  nombre d'arcs parcourus avant de trouver $S$, si elle est présente,
  est $n$ (le parcours doit s'achever sur un n{\oe{}}ud étiqueté comme
  n{\oe{}}ud final).  Le nombre d'arcs parcourus avant de déterminer
  que $S$ n'est pas présente dans le dictionnaire est $k\leqslant{}n$
  (au pire: 4 arcs parcourus avant de constater que le mot «\,abri\,»
  n'est pas dans le dictionnaire).}
\label{pppa}
\end{center}
\end{figure}

Lorsqu'on a affaire à des corpus que l'on a le temps de
pré-traiter «\,hors-ligne\,» pour construire un index, le
temps de recherche peut être considérablement diminué (ramené
à $O(n\times{}\log_2(k))$ dans le cas d'arbres binaires de
recherche équilibrés, par exemple), voire rendu indépendant
de la taille du corpus: c'est le cas lorsqu'on utilise un arbre
lexicographique. Dans ce cas, le temps de recherche ne dépend plus
que de la longueur de la chaîne recherchée ($O(n)$)\footnote{Il
  dépend aussi, secondairement, d'un autre facteur qui est le taux
  de branchement des arcs sortants à la sortie de chaque n{\oe{}}ud
  de l'arbre; ce dernier facteur dépend lui-même de deux autres
  facteurs, qui sont la taille de l'alphabet, et la
  «\,densité\,» de l'ensemble des mots dans le monoïde
  libre. On peut considérer que ce dernier paramètre est
  indirectement lié à la taille du corpus (plus le corpus est
  grand, plus il contient de vocables distincts), mais les études
  sur les lois de distribution des mots dans les langues
  \cite{Herdan1966} montrent que ce nombre croît de plus en plus
  lentement à mesure que la taille du corpus augmente (loi de
  Zipf-Mandelbrot).}.  Le principe de la recherche exacte dans un
arbre lexicographique est illustré dans la figure~\ref{pppa}.

Pour travailler avec la mesure de divergence entre chaînes $d_C$, qui
ne possède pas la propriété d'inégalité triangulaire, il est important
que l'élimination de branches de l'espace de recherche ne se fasse pas
avant d'avoir vérifié que l'on incluait les courts-circuits possibles.

En effet un couple de chaînes répertorié dans $\Gamma{}^0$ peut avoir
une distance court-circuitée plus faible que celle que l'on
obtiendrait en progressant de façon monotone dans les deux chaînes par
concaténation successive de symboles.  Imaginons par exemple que, pour
prendre en compte une faute d'orthographe possible, on inclue dans la
définition de $(\Gamma{}^0,c^0)$ le court-circuit:
$d_C($``occident''$,$``oxydant''$)=1,5$\footnote{Ceci n'est pas
  identique au fait de poser des courts-circuits hors-contexte:
  $d_C($``cc''$,$``x''$)=0,5$, $d_C($``i''$,$``y''$)=0,5$, et
  $d_C($``e''$,$``a''$)=0,5$, ce qui conduirait à des
  surgénéralisations abusives.}.  On a alors dans ce cas d'une part
$d_C($``occiden''$,$``oxydan''$)=4$, et d'autre part
$d_C($``occident''$,$``oxydant''$)=1,5$.  Si l'on se contentait de
progresser symbole après symbole dans la chaîne recherchée, on aurait
donc élagué le chemin conduisant à ``occident'' avant d'arriver au
dernier caractère.

La recherche approximative de chaînes, fondée sur le principe de
l'élagage d'arbre de recherche (élimination des branches lorsque la
valeur de $d_C$ dépasse un certain seuil) doit donc prendre en compte
les courts-circuits à partir de l'endroit où ils commencent, et non à
partir de l'endroit où ils se terminent.  L'absence de la propriété
d'inégalité triangulaire nous oblige à inclure dans l'algorithme un
mécanisme de contrôle explicite de l'absence de cycle (c'est le sens
du test «\,{\bf si} $\delta{}+c^1(G,H)<\delta{}'$\,»), mais il s'agit
d'une contrainte indépassable liée à la nature du problème, comme
l'illustre l'exemple du paragraphe précédent.

Cette question peut malgré tout être résolue avec
l'algorithme~\ref{rechapprox}.  Dans cet algorithme, on suppose qu'on
a accès directement à tous les couples de $\Gamma{}^1$ (tels que
définis plus haut, section~\ref{distance}), ainsi qu'aux valeurs de
$c^1$ qui leur correspondent.  La définition de l'algorithme est ainsi
simplifiée par le fait que la symétrie de la divergence entre chaîne
recherchée et correspondances approximatives partielles est encodée
dans la définition symétrique de $c^1$, de même que les cas
d'insertions (couples $(\epsilon{},{\mbox{``}}\sigma{}{\mbox{''}})$),
de suppressions (couples
$({\mbox{``}}\sigma{}{\mbox{''}},\epsilon{})$), ou de substitutions
(couples
$({\mbox{``}}\sigma{}{\mbox{''}},{\mbox{``}}\tau{}{\mbox{''}})$ où
$\tau{}\neq{}\sigma{}$).  Ceci n'interdit pas de ne stocker «\,en
dur\,», sous forme de table de valeurs, que l'un des triangles
diagonaux de $c^0$, et de déléguer à une fonction le calcul de $c^1$
(intégrant les couples de $\Gamma{}^1-\Gamma{}^0$ et les couples
symétriques).

On note $S[i:j]$ la sous-chaîne de $S$ commençant à la position $i$ et
se terminant à la position $j$ (p.ex., si $S={\mbox{``abricot''}}$,
$S[3:5]={\mbox{``ic''}}$), $S[:j]$ le préfixe de $S$ allant de la
position $0$ à la position $j$, et $S[i:]$ le suffixe de $S$ allant de
la position $i$ à la fin ($|S|$).

On définit également une fonction {\em prolongement} prenant en entrée
un pointeur vers un n{\oe{}}ud $\boldsymbol{n}$ de l'arbre
lexicographique et une chaîne $S\in{}\Sigma{}^*\cup{}\{\epsilon{}\}$,
et qui donne en retour un autre pointeur (éventuellement NULL) vers un
n{\oe{}}ud de l'arbre. Cette fonction est définie de telle sorte que
si un chemin existe entre $\boldsymbol{n}$ et un autre n{\oe{}}ud
$\boldsymbol{n'}$ de l'arbre en passant successivement par tous les
symboles de $S$, un par un, alors ${\mbox{\em
    prolongement}}(\boldsymbol{n},S)=\boldsymbol{n'}$; sinon,
${\mbox{\em prolongement}}(\boldsymbol{n},S)={\mbox{NULL}}$.

\setlength{\algomargin}{0px}

\begin{algorithm}
\label{rechapprox}
\caption{Recherche approximative de chaîne}
\SetKwInput{Variables}{Variables}
\DontPrintSemicolon
\Entree{\\
$A$: arbre lexicographique de n{\oe{}}ud racine $\boldsymbol{n_0}$ (dictionnaire encodé sous forme d'arbre lexicographique)\\
$S\in{}\Sigma{}^*$: chaîne de longueur $n$~; $s\in{}\mathbb{R}^+$: seuil de distance toléré}
\Sortie{\\
$R$: ensemble de couples $(T,d_C(S,T))\in{}\Sigma{}^*\times{}\mathbb{R}^+$ (chaînes $T$ du dictionnaire avec leur distance $d_C$ à $S$)}
\Variables{\\
$V$: pile de triplets $(i,\boldsymbol{n},\delta{})$ où $i\leqslant{}n\in{}\mathbb{N}$ est un index sur $S$, $\boldsymbol{n}$ un n{\oe{}}ud de $A$, et $\delta{}\in{}\mathbb{R}^+$ une mesure de la distance $d_C$ entre $S[:i]$ et le préfixe encodé par le n{\oe{}}ud $\boldsymbol{n}$\\
$t$ triplet de $V$~; $G,H$ blocs (sous-chaînes) telles que $(G,H)\in{}\Gamma{}^1$\\
$g\in{}\mathbb{N}$ nombre d'extensions possibles trouvées pour un préfixe de $S$, à chaque itération
}
\Deb{
  $V \leftarrow{} \{(0,\boldsymbol{n_0},0)\}$\;
  \Repeter{g=0}{
    $g \leftarrow{} 0$\;
    \PourTous{$t=(i,\boldsymbol{n},\delta{})\in{}V$}{
      \PourTous{$(G,H)\in{}\Gamma{}^1$}{
        $\boldsymbol{n'}={\mbox{\em prolongement}}(\boldsymbol{n},H)$\;
        \Si{$\boldsymbol{n'}\neq{\mbox{NULL}}$}{
          \Si{$S[i:i+|G|]=G$ {\bf et} $\delta{}+c^1(G,H)<s$}{
            $t'=(i+|G|,\boldsymbol{n'},\delta{}+c^1(G,H))$\;
            \uSi{$\exists{}(i+|G|,\boldsymbol{n'},\delta{}')\in{}V$}{
              \Si{$\delta{}+c^1(G,H)<\delta{}'$}{
                remplacer $(i+|G|,\boldsymbol{n'},\delta{}')$ par $t'$ dans $V$\;
                $g \leftarrow{} g+1$
              }
            }
            \Sinon{
              $V \leftarrow{} V \cup{} t'$\;
              $g \leftarrow{} g+1$
            }
          }
        }
      }
      \Si{$i<|S|$}{
        retirer $t$ de $V$\;
      }
    }
  }
  $R \leftarrow{} \{\}$\;
  \PourTous{$t=(i,\boldsymbol{n},\delta{})\in{}V$}{
    \Si{i=n}{
      $T \leftarrow{}$ chaîne encodée par le n{\oe{}}ud $\boldsymbol{n}$ de $A$\;
      $R \leftarrow{} R\cup{}\{(T,\delta{}/n)\}$\;
    }
  }
  \Retour{$R$}\;
}
\end{algorithm}

\subsection{Considérations sur la complexité}

Cet algorithme ne prend pas en compte la possibilité
d'altérations {\em récursives} sur $S$.  La notion de
récursivité des altérations a été exposée par
\cite{ShapiraStorer2011}: une séquence d'opérations est
récursive si elle modifie des blocs qui n'étaient pas
eux-mêmes présents dans la chaîne d'origine.  Nous postulons
que nous n'avons pas besoin de prendre en compte des cas de ce type.
Nous devrions le faire si nous souhaitions retrouver, par exemple, la
chaîne ``pâte'' à partir de la chaîne ``patoute'', en
faisant l'hypothèse que ``pâte'' a d'abord été
altérée par l'utilisateur en ``patte'', puis que le ``tt'' dans
``patte'' était lui-même une abréviation de ``tout''.  Il
nous semble que ces cas de figure ne correspondent pas à des
problèmes réels dans le cas où les séquences de symboles
traités sont des chaînes de caractères produites dans une
langue humaine\footnote{Ce n'est pas le cas lorsque l'on
  s'intéresse à la similarité de séquences d'ADN ayant pu
  subir plusieurs altérations successives~--- contexte pour lequel
  cette récursivité a été définie.}.  Au
pire, il peut y avoir des cas de figure où une chaîne peut
être exposée, dans le cadre de son processus de
génération, à plusieurs sources d'altération possibles de
nature différente (p.ex. un texte engendré par un logiciel d'OCR
puis réédité par un relecteur utilisant un clavier);
cependant, nous partons du principe que dans ces cas-là, toute
l'information concernant ces différentes catégories
d'altérations est contenue dans la définition de $\Gamma{}^0$ et
$c^0$.

Considérant que nous ne cherchons pas à prendre en compte des
altérations récursives, et que nous limitons par ailleurs la
liste des courts-circuits à un ensemble limité, l'algorithme
décrit ci-dessus a des propriétés, en termes de
complexité, qui le maintiennent dans la norme des algorithmes de
recherche définis sur des opérations élémentaires portant
sur des caractères individuels.

Comme eux, son point faible est qu'il est extrêmement sensible à
deux facteurs, qui ont un impact sur la quantité maximale de
triplets $t$ (n{\oe{}}uds correspondant à une correspondance
approximative avec un préfixe de $S$) qui doivent être gardés
en mémoire dans $V$ au cours de l'exécution de l'algorithme.

Le premier, que nous pouvons noter $\beta{}$, est le taux de
branchement (le nombre d'arcs sortants par n{\oe{}}ud) dans l'arbre
lexicographique.  Le second est le rapport entre le seuil d'élagage
($s$) et la distance minimale non-nulle $\varepsilon{}$ que l'on peut
attribuer à un court-circuit élémentaire dans $\Gamma{}^0$.
Plus ce rapport est grand, et plus on peut, en théorie, être
amené à garder en mémoire un grand nombre de n{\oe{}}uds
intermédiaires dans $V$.  Si nous imaginons par exemple que l'on
ait décidé d'imputer un coût de $0,25$ à l'insertion d'un
caractère $a$ ($c^0({\mbox{``}}a{\mbox{''}},\epsilon{})=0$), et que
le seuil $s$ est de $1$, alors l'algorithme pourra être amené à
garder des chemins contenant jusqu'à quatre arcs $a$ inutiles,
avant de les élaguer.

Ainsi, dans le pire des cas (si tous les symboles ont un court-circuit
de coût $\varepsilon{}$ vers la chaîne vide), $V$ doit contenir
des n{\oe{}}uds de $A$ qui peuvent avoir été atteint par des
chemins comportant $s/\varepsilon{}$ arcs inutiles, ce qui implique
potentiellement $\beta{}^{s/\varepsilon{}}$ branchements inutiles.

On note par ailleurs $\kappa{}$ le nombre de courts-circuits de
$\Gamma{}^0$ définis sur un couple de blocs $(G,H)$ dont au moins
un est de longueur strictement supérieure à $1$ (courts-circuits
définis sur plus d'un caractère à la fois).  On sait que
$\Gamma{}^1$, par construction, contient tous ces couples-là, plus
tous les couples de blocs de longueur $0$ ou $1$ (ceux de
$\Sigma{}\cup{}\{\epsilon{}\}\times{}\Sigma{}\cup{}\{\epsilon{}\}-\{(\epsilon{},\epsilon{})\}$),
donc en tout $(|\Sigma{}|+1)^2+\kappa{}-1$ couples.

Chaque itération devant passer en revue l'ensemble des couples de
$\Gamma{}^1$ pour chacun des triplets non-élagués de $V$, on voit que
l'algorithme reste nominalement linéaire ($O(n)$ en fonction de la
longueur $n$ de la chaîne recherchée), mais avec un facteur constant
$(\beta{}^{s/\varepsilon{}})\times{}((|\Sigma{}|+1)^2+\kappa{}-1)$ qui
peut devenir énorme, selon la manière dont le seuil d'élagage $s$ et
le palier minimal de distance pour $c^0$, $\varepsilon{}$, sont
définis.

\section{Conclusion et perspectives}
\label{conclusion}

Nous avons présenté un algorithme de recherche approximative
dans un dictionnaire, capable de trouver l'ensemble des entrées
contenues dans une fourchette de distance déterminée d'une forme
non-normalisée.  Cet algorithme se distingue de travaux
antérieurs en ce qu'il prend en compte, pour la définition de
cette distance, d'une mesure qui peut être adaptée avec
souplesse à un corpus donné, car elle offre la possibilité de
définir des altérations à l'échelle de blocs de symboles
(«\,courts-circuits\,») et pas seulement entre symboles pris un
par un.

La question qui n'a pas été abordée ici (elle fait l'objet
d'un travail encore en cours) est celle de l'acquisition d'une telle
distance par apprentissage automatique sur le corpus.  La distance
définie ici n'est pas une véritable distance, au sens de
fonction définie positive définissant un espace métrique, car
elle ne possède pas la propriété d'inégalité
triangulaire; cependant, elle est symétrique et séparable, et
devrait donc pouvoir être acquise par des procédures
d'apprentissage semi-supervisé à partir de jeux d'exemples et de
corpus d'entraînements.

\bibliography{biblio-vaillant-coria2015}

\end{document}